\pdfoutput=1

\documentclass[11pt]{article}
\usepackage[]{authblk}

\usepackage[]{emnlp2021}
\usepackage{times}
\usepackage{latexsym}

\usepackage{url}
\usepackage{svg}
\usepackage{comment}
\graphicspath{{Figures/}}
\usepackage[title]{appendix}
\usepackage{enumitem}

\usepackage{tablefootnote}
\usepackage{xcolor}
\usepackage{CJKutf8}
\usepackage{multirow}
\usepackage{soul}
\usepackage{booktabs}

\newcommand{\nop}[1]{}

\usepackage{pbox}
\usepackage{xspace}
\newcommand{\cough}{\texttt{COUGH}\xspace}

\usepackage[T1]{fontenc}

\usepackage[utf8]{inputenc}

\usepackage{microtype}

\title{\cough: A Challenge Dataset and Models for COVID-19 FAQ Retrieval}

\author[1,2,\thanks{\, Work was done when the first two authors were at OSU.}]{Xinliang Frederick Zhang}
\author[1,3,*]{Heming Sun}
\author[1]{Xiang Yue}
\author[4]{Simon Lin}
\author[1]{Huan Sun}

\affil[1]{The Ohio State University}
\affil[2]{University of Michigan}
\affil[3]{University of Southern California}
\affil[4]{Abigail Wexner Research Institute at Nationwide Children’s Hospital}
\affil[ ]{\{\texttt{zhang.9975,sun.2164,yue.149,sun.397\}@osu.edu,}}
\affil[ ]{\texttt{Simon.Lin@nationwidechildrens.org}}

\begin{document}
\maketitle

\begin{abstract}
We present a large, challenging dataset, \cough,  for COVID-19 FAQ retrieval. Similar to a standard FAQ dataset, \cough consists of three parts: FAQ Bank, Query Bank and  Relevance Set. The FAQ Bank contains $\sim$16K FAQ items scraped from 55 credible websites (e.g., CDC and WHO). For evaluation, we introduce Query Bank and  Relevance Set, where the former contains 1,236 human-paraphrased queries while the latter contains $\sim$32 human-annotated FAQ items for each query. We analyze \cough by testing different FAQ retrieval models built on top of BM25 and BERT, among which the best model achieves 48.8 under P@5, indicating a great challenge presented by \cough and encouraging future research for further improvement. Our \cough dataset is available at \url{https://github.com/sunlab-osu/covid-faq}.
\end{abstract}
\section{Introduction}
 
Many institutional websites today maintain an FAQ page to help users find relevant information for commonly asked questions. The FAQ retrieval task is defined as ranking FAQ items $\{(q_i,a_i)\}$\footnote{$q$ and $a$ are question and answer fields in an FAQ item.} from a collection given a user query $Q$ \citep{Karan2016}. In contrast to common Information Retrieval (IR), FAQ retrieval often introduces 3 new challenges: {1) brevity of FAQ texts in comparison with IR documents;}
2) need for topic-specific knowledge; 3) usage of the new question field in FAQ items \citep{Karan2016,FAQ-Sakata}. However, FAQ retrieval is under-studied compared with other IR applications such as open-domain QA \citep{QA2020Chen}.

{In this work, we specifically study FAQ retrieval for COVID-19, a contagious and fatal pandemic which is still evolving on a daily basis. Many websites like CDC and WHO provide quality information on COVID-19 and update FAQ pages regularly.}

\begin{figure}[t]
    \centering
    \includegraphics[width=0.9\linewidth]{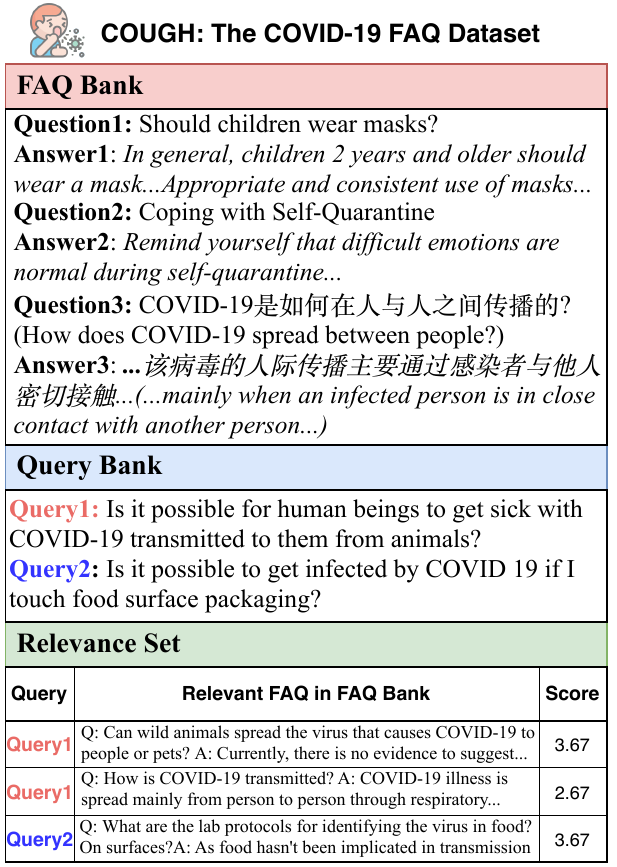}
    \vspace{-5pt}
    \caption{Examples from the \cough dataset.}
    \label{fig:intro_eg}
    \vspace{-10pt}
\end{figure}
\begin{table*}[t]
\centering
\resizebox{\linewidth}{!}{%
\begin{tabular}{lrrrrrr}
 & \pbox{20cm}{\textbf{FAQIR}\\(\citeauthor{Karan2016})}  & \pbox{20cm}{\textbf{StackFAQ}\\(\citeauthor{Karan2018})} & \pbox{20cm}{\textbf{LocalGov}\\(\citeauthor{FAQ-Sakata})} & \textbf{\citeauthor{covid-Sun}}  & \textbf{\citeauthor{covid-Poliak}}  & \textbf{\cough (ours)} \\ \hline
Domain & Yahoo! & StackExachange & Government & COVID-19 & COVID-19 & COVID-19 \\
\# of FAQs & 4,313 & 719 & 1,786 & 690 & 2,115 & 15,919 \\
\# of Queries (Q) & 1,233 & 1,249 & 784 & 6,495* & 24,240* & 1,236 \\
\# of annotations per Q & 8.22 & Not Applicable & $<$10 & 5 & 5 & 32.17 \\ \hline
Query Length & 7.30 & 13.84 & ** & ** & 6.87 & 12.97 \\
FAQ-query Length & 12.30 & 10.39 & ** & ** & 8.73 & 13.00 \\
FAQ-answer Length & 33.00 & 76.54 & ** & ** & 76.71 & 113.58 \\ \hline
Language & English & English & Japanese & English & Multi-lingual & Multi-lingual \\
\# of sources & 1 & 1 & 1 & 12 & 34 & 55 \\ \hline
\end{tabular}
}
\vspace{-5pt}
\caption{Comparison of \cough with representative counterparts. *: Extracted from existing resources (e.g., COVID-19 Twitter dataset \cite{Chen_2020}).  **: Not Applicable, either not in English or not publicly available.  }
\vspace{-10pt}
\label{tbl:compare}
\end{table*}

To gain better insights into FAQ retrieval research and advance COVID-19 information search, we present an FAQ dataset, \cough\footnote{Adapted from ``CoF'' that stands for \textbf{CO}VID \textbf{F}AQ.}, consisting of FAQ Bank, Query Bank and Relevance Set, following the standard of constructing an FAQ dataset \cite{Manning-IR}. 
The FAQ  Bank  contains 15919  FAQ  items scraped from 55 authoritative institutional websites (see a full list in Table \ref{tbl:index_English} and \ref{tbl:index_Non-English}). \cough covers a wide range of topics on COVID-19, from general information about the virus to specific COVID-related instructions for a healthy diet. For evaluation, we further construct Query  Bank and Relevance Set, including 1,236 crowd-sourced queries and their relevance to a set of FAQ items judged by annotators. 
Examples from \cough are shown in Figure \ref{fig:intro_eg}.

Our dataset poses several new challenges (e.g., answer fields are longer and noisier, and harder to match, than question fields) to existing \nop{FAQ retrieval models} methods. The diversity of FAQ items, reflected in varying query forms and lengths as well as in narrative styles, also contributes to these challenges.

{The contribution of this work is two-fold. First, we construct a challenging dataset \cough to aid the development of COVID-19 FAQ retrieval models. Second, we evaluate various FAQ retrieval models across different settings, explore their limitations, and encourage future work along this line.} 

\section{Related Work}
\label{related}
\noindent \textbf{COVID-19 \& FAQ Datasets.}
Since the outbreak of COVID-19, the community has witnessed many datasets released to advance the research of COVID-19. For example, CORD-19 \cite{wang2020cord19}, CODA-19 \cite{CODA-19-COV}, COVID-Q \cite{wei2020people}, Weibo-Cov \cite{Weibo-COV}, and Twitter dataset \cite{Chen_2020}. All of them aim to aggregate resources to combat COVID-19.

The most related works to ours are \citet{covid-Sun} and \citet{covid-Poliak}, both of which constructed a collection of COVID-19 FAQs by scraping authoritative websites. 
However, the dataset in the former work is not available yet and the latter work does not evaluate models on their dataset, and there is still a great need to understand how existing models would perform on the COVID-19 FAQ retrieval task. In the open domain, several FAQ datasets appeared recently, such as FAQIR \cite{Karan2016}, StackFAQ \cite{Karan2018} and LocalGov \cite{FAQ-Sakata}.
Unfortunately, as shown in Table \ref{tbl:compare}, the scale of existing FAQ datasets is too small, and answer lengths are much lower than those in \cough, which may not characterize the difficulty of FAQ retrieval tasks in real-world scenarios. Moreover, in contrast to all prior datasets, \cough covers multiple query forms (e.g., question and query string forms) and has many annotated FAQs for each user query, whereas queries in existing FAQ datasets are limited to the question form and have much fewer annotations.

\noindent \textbf{FAQ Retrieval Methods.}
FAQ retrieval focuses on retrieving the most-matched FAQ items given a user query \cite{Karan2018}. 
Many earlier works, e.g., FAQ FINDER \cite{Burke-1997}, query expansion \cite{KIM2006650} and BM25 \cite{INR-019}, resorted to traditional IR techniques by leveraging lexical mapping and/or semantic similarity. In the deep learning era, many studies show that Neural Networks are useful for FAQ retrieval as they are good at learning the semantic relevance between queries and FAQ items. 
Along this line, \citet{Karan2016} adopted Convolution Neural Networks, \citet{Gupta-LSTM} utilized LSTM, and \citet{FAQ-Sakata} leveraged an ensemble of TSUBAKI and BERT. Recently, \citet{Unsupervised-FAQ} explored learning to rank without requiring manual annotations. 
\section{Dataset Construction\footnote{We provide detailed
annotation protocols in Appendix~\ref{protocols}.}}
\label{construction}

\subsection{FAQ Bank Construction}

We developed scrapers\footnote{Scrapers are released together with \cough to keep FAQ Bank up-to-date.} adapted from \citet{covid-Poliak}, and add special features to \cough dataset.

\noindent\textbf{Web scraping:} We collect FAQ items from authoritative international organizations, state governments and other credible websites including reliable encyclopedias and medical forums. 
Moreover, we scrape three types of FAQs: question (i.e., an interrogative statement), query string (i.e., a string of words to elicit information) and forum (FAQs scrapped from medical forums) forms. 
Inspired by \citet{Manning-IR}, we loosen the constraint that queries must be in question form since we want to study a more generic and challenging problem. We also scrape 6,768 non-English FAQs to increase language diversity. Overall, we scraped 15,919 FAQ items covering all three forms and 19 languages. 

\subsection{Query Bank Construction}
\label{userQueryBankConstruction}

Following \citet{Manning-IR, Karan2016}, we do not crowdsource queries from scratch, but instead ask annotators to paraphrase our provided query templates. That way, we ensure that 1) collected queries are pertinent to COVID-19; 2) collected queries are not too simple; 3) the chance of getting similar user queries is reduced. 

\noindent\textbf{Phase 1: Query Template Creation}: We sample 5\% of FAQ items from each English non-forum source\footnote{Each source contributes at least one item to ensure wide topic coverage and similar sampled FAQ items are removed.} and use the question part as the \textit{template}. For example, the templates of the two paraphrased queries in Figure~\ref{fig:intro_eg} are ``Can humans become infected with the COVID-19 from an animal source?'' and ``Can I get sick with COVID-19 from touching food, the food packaging, or food contact surfaces, if the coronavirus was present on it?''.

\noindent\textbf{Phase 2: Paraphrasing for Queries}: In this phase, each annotator is expected to give three paraphrases for each query template. Besides providing shallow parapharases (e.g., word substitution), annotators are encouraged to give deep paraphrases (i.e.,  grammatically different but semantically similar/same) to simulate the noisy and diverse environment in real scenarios. In the end, we obtain 1,236 human-paraphrased user queries.

\subsection{Relevance Set Construction}
\label{AnnotatedRelevanceSetConstruction}
\noindent\textbf{Phase 1: Initial Candidate Pool Construction}: For each user query, as suggested by previous work \cite{Manning-IR,Karan2016, FAQ-Sakata}, we run 4 models (see Section \ref{Methods}), BM25 (Q-q), BM25 (Q-q+a), BERT (Q-q), and BERT (Q-a) fine-tuned on \cough, to instantiate a candidate FAQ pool.  Each model complements the others and contributes its top-10 relevant FAQ items. We then take the union to remove duplicates, giving an average pool size of 32.2.

\noindent\textbf{Phase 2: Human Annotation}: Each annotator gives each $\langle$Query, FAQ item$\rangle$ tuple a score based on the annotation scheme (i.e., 4/Matched, 3/Useful, 2/Useless and 1/Non-relevant)\footnote{Table~\ref{task2annotation} details the meaning of these four scores.} adapted from \citet{Karan2016, FAQ-Sakata}. In order to alleviate the annotation bias, each tuple has at least 3 annotations. In the finalized Set, we keep all raw scores and include: 1) mean of annotations; {2) four suggested aggregation schemes to obtain binary labels (as detailed in Appendix~\ref{labeling}). Users of \cough can also try other aggregation measures.} 

Among 1,236 user queries, there are 35 ``unanswerable" queries that have no associated positive FAQ item. 

\begin{table}[t]
\centering
\resizebox{\linewidth}{!}{%
\begin{tabular}{lrrrr}
\hline
 & Type &Number & Q-Length & A-length\\ \hline
 \hline
 \multirow{3}*{\# English}
    & Question & 4,978 & 14.64 & 123.89  \\ 
    & Query String  & 2,139 & 9.18 & 89.60  \\ 
    & Forum & 2,034 & 147.46 & 90.49  \\ 
\hline

  \multirow{2}*{\# Non-English}
     &  Question &3,396  & - & - \\ 
     & Query String  &3,372 & - & -  \\
     
\hline
\# Total & - &15,919  & - & - \\ \hline
\end{tabular}
}
\vspace{-5pt}
\caption{Basic statistics of FAQ bank in \cough.}
\vspace{-10pt}
\label{tbl:data}
\end{table}

\section{Dataset Analysis}
\label{data-analysis}

Besides the generic goal of large size, diversity, and low noise, \cough features 5 additional aspects. 

\noindent \textbf{Varying Query Forms:}
As indicated in Table \ref{tbl:data}, there are multiple query forms. In evaluation, we include both question {(Question1 and 3 in Figure~\ref{fig:intro_eg}) }and query string {(Question2 in Figure~\ref{fig:intro_eg})} forms. These two distinct forms are  different in terms of query format (interrogative v.s. declarative), average answer length (123.89 v.s. 89.60) and topics. Question form is usually related to general information about the virus while query string form is often searching for more specific instructions concerning COVID-19 (e.g., healthy diet during pandemic).

\noindent \textbf{Answer Nature:}
Table \ref{tbl:compare} shows the answer fields in \cough are much longer than those in any prior dataset. We also observe that answers might contain some contents which are not directly pertinent to the query, partially resulting in the long length nature. For example, in \cough, the answer to a query ``What is novel coronavirus" contains extra information about comparisons with other viruses. Such lengthy and noisy nature of answers manifest the difficulty of FAQ retrieval in real scenarios.

\noindent \textbf{Language Correctness in Query Bank:}
Most queries in our Query Bank are properly spelled and grammatically correct, so we can prioritize investigating the model performance under a less noisy setting. Furthermore, our dataset can support a controlled study on the impact of spelling and grammatical errors: One can simulate various kinds of spelling and grammatical errors and inject them in a controlled manner into the Query Bank and systematically evaluate how the model performance changes under different levels of noises.

\noindent \textbf{Large-scale Relevance Annotation:} 
Many existing FAQ datasets overlooked annotation scale (Table \ref{tbl:compare}); yet, that would hurt the  evaluation reliability since many true positive $\langle$Query, FAQ item$\rangle$ tuples were omitted. Following \citet{Manning-IR}, for each user query, we constructed a large-scale candidate pool to reduce the chance of missing true positive tuples. The annotation procedure yielded 39760 annotated  tuples, each of which is annotated by at least 3 people to reduce annotation bias.

\noindent \textbf{Multilinguality:}
\begin{figure}[t]
  \vspace{-5pt}
  \centering
  \includegraphics[width=0.5\textwidth]{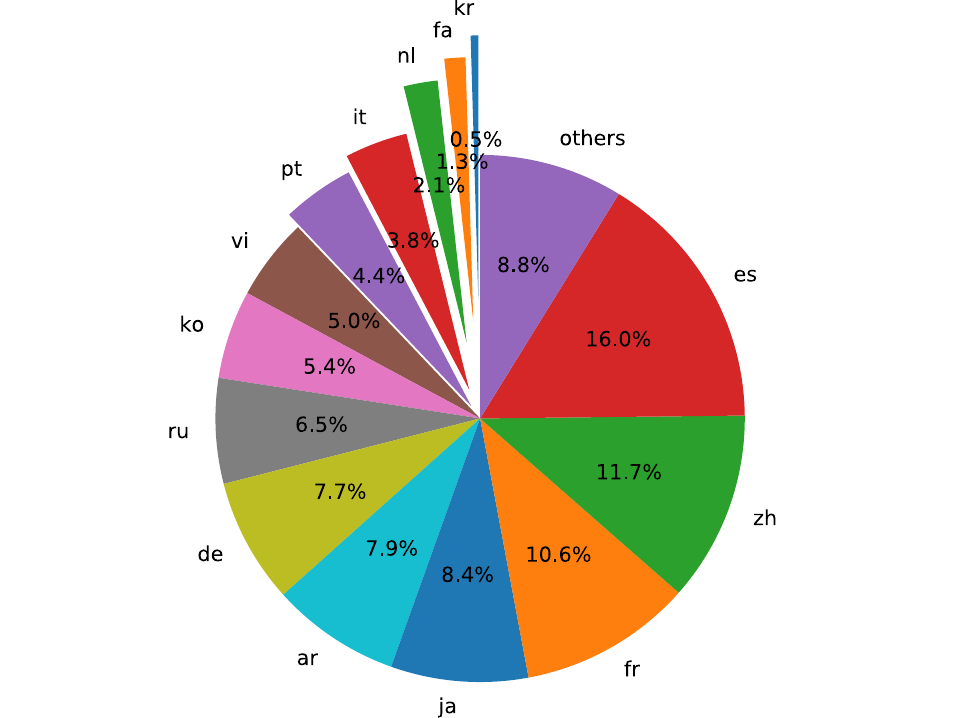}
  \vspace{-10pt}
  \caption{Language distribution for non-English FAQ items.}
  \vspace{-10pt}
  \label{fig:language}
\end{figure}
\cough includes 6768 FAQ items covering 18 non-English languages, and statistics of non-English items can be found in Table~\ref{tbl:data}. Figure \ref{fig:language} shows the language distribution (excluding English) of FAQ items in \cough dataset. Like English FAQ items, non-English FAQ items are also presented in both question and query string forms. 
The detailed breakdown of non-English portion by sources and languages is shown in Table \ref{tbl:index_Non-English}.

However, due to budget limit, we did not proceed to the annotation phase for non-English data, so there is no non-English human-paraphrased user query or relevance judgement.

\noindent\textbf{Annotation Quality:} We discard low-quality  paraphrased queries ($\sim$24\%) and relevance annotations ($\sim$11\%). Further, we show that $\sim$74\% of annotated tuples have high agreements where multiple people vote for the same relevance class. More details of quality checking can be found in Section~\ref{ethical_dataset}.

\section{Experiments}

\subsection{Experimental Setup}

In this work, we focus on \textit{unsupervised} sparse and dense retrievers and discuss their limitations. Supervised learning is less popular for this task since it's too costly to collect a large-scale Query Bank and its associated relevance judgement \citep{ FAQ-Sakata,Unsupervised-FAQ}. Further, there are 3 configurable modes, Q-q, Q-a and Q-q+a, where a user query $Q$ can be learned to match with question $q$, answer $a$ or the concatenation $q$$+$$a$.

\subsection{Methods}
\label{Methods}

\noindent (1) \textbf{BM25} is a nonlinear combination of term frequency, document frequency and document length.

\vspace{1mm}
\noindent (2) \textbf{BERT} \cite{devlin2019bert} is a pretrained language model. 
We use its variant, Sentence-BERT \cite{sentence-bert}, to encode $Q$, $q$ and $a$ separately to generate sentence representations.

\noindent {\textbf{Fine-tuning:} Similar to \citet{henderson2017efficient, DPR}, we leverage in-batch negatives\footnote{In a batch, $(q_i,p_j)$ is assumed as negative pair if $i\neq j$.} to fine-tune BERT on FAQ bank.
For Q-q mode, we use GPT-2 \citep{radford2019language} to generate synthetic questions to match with Q.  For Q-a mode, an FAQ item $(q,a)$ itself is a positive pair
}

\noindent {\textbf{Re-rank:} In Q-a mode, answers are quite long, so the importance of selecting most-related spans from relevant answers to catch the nuance is amplified. As detailed in \citet{sentence-bert,Humeau-encoder}, cross-encoder
can perform self-attention between query and answer, resulting in a richer extraction mechanism.  We re-rank\footnote{Directly applying cross-encoder is not efficient and yields inferior results in our preliminary experiments.} top-10 retrieved answers using cross-encoder BERT.}

\vspace{1mm}

\noindent {(3) \textbf{CombSum} \cite{Unsupervised-FAQ}
first computes three matching scores between the user query and FAQ items via BM25 (Q-q), BERT (Q-q)
and fine-tuned BERT (Q-a) models. Then, the three scores are normalized and combined by averaging. We also evaluate with no BERT (Q-a) included.
}

\subsection{Evaluation}
\label{eval}
\noindent \textbf{Evaluation Setting:} For the scope of this work, we only evaluate on 1,201 ``answerable'' English non-forum FAQ items, and leave the ``unanswerable'', non-English and forum ones for future research as great challenges have been observed under current setting. However, we encourage investigators to utilize those three categories for other potential applications (e.g., multi-lingual IR, transfer learning in IR). 

\noindent \textbf{Evaluation Metrics:} Following previous work \citep{Manning-IR,Karan2016, Karan2018,FAQ-Sakata,Unsupervised-FAQ}, we adopt P@1 (Precision), P@5, MAP@100 (Mean Average Precision), MRR (Mean Reciprocal Rank) and nDCG@5 (Normalized Discounted Cumulative Gain) as evaluation metrics.

\section{Analysis}
\begin{table}[t]
\centering
\resizebox{\linewidth}{!}{
\begin{tabular}{lrrrrr}
\hline
Method & P@1 & P@5 & MAP & MRR & nDCG \\ \hline
BM25 (Q-q) & 60.4 & 43.7 & 28.2 & 73.0 & 76.7 \\
BM25 (Q-a) & 33.4 & 25.6 & 16.2 & 47.4 & 46.4 \\
BM25 (Q-q+a) & 56.9 & 41.3 & 28.5 & 70.0 & 72.6 \\ \hline
BERT (Q-q) & 63.8 & 46.0 & 27.1 & 75.7 & \textbf{78.6} \\
\, + fine-tune on pesudo Q-q & 64.9 & 40.9 & 27.5 & 75.1 & 63.0 \\
BERT (Q-a) & 13.5 & 9.6 & 4.8 & 24.1 & 16.7 \\
\, + fine-tune on FAQ Bank & 52.0 & 37.1 & 25.8 & 66.0 & 56.4 \\
\, \, \, + re-rank & 52.1 & 38.4 & 26.4 & 66.3 & 57.8 \\ \hline
CombSum & \textbf{69.7} & \textbf{48.8} & \textbf{37.3} & \textbf{80.2} & 74.7 \\
\, - fine-tuned BERT (Q-a) & 65.4 & 45.8 & 31.5 & 77.2 & 75.2 \\ 
\hline
\end{tabular}
}
\vspace{-5pt}
\caption{Evaluation on \cough.}
\vspace{-10pt}
\label{tbl:extra}
\end{table}
\noindent \textbf{Quantitative Analysis.} Models' results, based on aggregation scheme A: annotated tuples with mean score $\geq$ 3 are positives, 

are listed in Table \ref{tbl:extra}. Results under other schemes are in appendix~\ref{labeling_results}.

The current best P@5 and MAP, 48.8 and 37.3, are not satisfying, showing a large room for improvement, {confirming that \cough is challenging.}

We observe that Q-q mode consistently performs better than Q-a mode.
{This is} because question fields are more similar to user queries than answer fields. As shown in Section \ref{data-analysis}, the answer nature (lengthy and noisy), albeit well characterizes the FAQ retrieval task in real scenarios, does bring up a great challenge. Utilizing the cross-encoder for re-ranking can yield better results since it can select query-aware features from answers. This is a possible step towards  handling long and noisy answers better.

{We also find that fine-tuning under the Q-a mode can improve the performance (e.g., from 9.6 to 37.1 under P@5), but might hurt it under the Q-q mode due to noises introduced by synthetic queries. Moreover, the best overall performances are achieved by BERT (Q-q) and CombSum, which are in line with \citet{Unsupervised-FAQ}. }
{However, CombSum without fine-tuned BERT (Q-a) performs worse than the original one. It indicates that answer fields can serve as supplementary resources for the missing information in the question field.}

\noindent \textbf{Qualitative Analysis.} To understand fine-tuned BERT (Q-q) better, we conduct case analyses in Table \ref{tbl:error} to show its major types of errors, hoping to further improve it in the future. 
Currently, fine-tuned BERT (Q-q) suffers from the following issues: 1) biased towards responses with similar texts (e.g., ``antibody tests'' and ``antibody testing''); 2) fails to capture the semantic similarities under complex environments (e.g., pragmatic reasoning is required to understand that ``limited abiltity'' indicates results are not accurate for diagnosing COVID-19). 

Interesting future work includes: 1) handling long and noisy answer fields, e.g., via {salient span selection}; 2) further improving semantic understanding or reasoning skills, {beyond lexical match}.

\begin{table}[t]
\resizebox{\linewidth}{!}{%
\begin{tabular}{p{1.45\linewidth}l}
\hline
\textbf{Query:} What research is being done on antibody tests and their accuracy?  \\
\textbf{FAQ item:} Q: What is antibody testing? How do I get a COVID-19 antibody test? A: CDC and partners are investigating to determine if you can get sick with COVID-19 more than once ...\\
\hline
\textbf{Gold label:} \texttt{Negative} \texttt{[}useful, useless, useless\texttt{]}\\ 
\textbf{Predicted rank:} \texttt{3}\\ \hline \hline
\textbf{Query:} Are COVID-19 antibody tests accurate?  \\
\textbf{FAQ item:}  Q: Should I be tested with an antibody (serology) test for COVID-19? A: ... Antibody tests have limited ability to diagnose COVID-19 and should not be used alone to diagnose COVID-19 ... \\\hline
\textbf{Gold label:} \texttt{Positive} \texttt{[}useful, useful, matched\texttt{]} \\ \textbf{Predicted rank:} \texttt{26} \\ \hline
\end{tabular}
}
\vspace{-5pt}
\caption{Case analyses with fine-tuned BERT (Q-q).  Human annotations are inside \texttt{[]}.}
\label{tbl:error}
\vspace{-10pt}
\end{table}

\section{Conclusion}
In this paper, we introduce \cough, a large challenging dataset for COVID-19 FAQ retrieval. \cough features varying query forms, long and noisy answers, and multilinguality. \cough also serves as a better evaluation benchmark since it has quality larger-scale relevance annotations. {We discuss the limitations of current FAQ retrieval models via comprehensive experiments, and encourage future research to further improve FAQ retrieval.}

\section*{Acknowledgments}
We thank our hired AMT workers for their annotations. We thank all anonymous reviewers for their helpful comments. We thank Emmett Jesrani for revising an earlier version of the paper. This research was sponsored in part by the Patient-Centered Outcomes Research Institute Funding ME-2017C1-6413, the Army Research Office under cooperative agreements W911NF-17-1-0412, NSF Grant IIS1815674, NSF CAREER 1942980, and Ohio Supercomputer Center \cite{OhioSupercomputerCenter1987}. The views and conclusions contained herein are those of the authors and should not be interpreted as representing the official policies, either expressed or implied, of
the Army Research Office or the U.S. Government. The U.S. Government is authorized to reproduce and distribute reprints for Government purposes notwithstanding any copyright notice herein.
\begin{table*}[t]
\resizebox{\linewidth}{!}{%
\begin{tabular}{lllrrl} \hline
 & \textbf{Task} & \textbf{Details} & \textbf{Base Cost} & \textbf{\pbox{20cm}{Base  Cost\\ per Unit}} & \textbf{\pbox{20cm}{Cognitive \\Complexity}} \\ \hline
\multicolumn{6}{c}{\textbf{User Query Bank Construction}} \\ \hline
\multirow{2}{*}{Reference} & QA annotation & Identify 3 QA pairs (write questions and then find answers). & 24 & 8 & High \\
 & \pbox{20cm}{Question annotation \\ on an audio clip}  & \pbox{20cm}{Identify 3 Questions (write questions and then find answers), each of \\which has additional requirements (e.g., originality, creativeness).}     & 30 & 10 & \pbox{20cm}{Extremely \\ high}\\ \hline
Ours* & Paraphrase Queries & Give 3 paraphrases for the original query template. & 12 & 4 & Medium \\ \hline
\multicolumn{6}{c}{\textbf{Annotated Relevance Set Construction}} \\ \hline
\multicolumn{1}{c}{\multirow{3}{*}{Reference}} & Image labeling & Locate 5 required objects in a given image. & 7 & 1.4 & Medium \\
\multicolumn{1}{c}{} & Website class identification & Identify the type of niche of a twitter account. Select from 6 classes. & 2 & 2 & Low \\
\multicolumn{1}{c}{} & Identify an item & Given an image, fill out a form with 6 required fields. & 9 & 1.5 & Medium \\ \hline
Ours* & Relevance judgements & Identify the relevance.  Select   from 4 classes. & 2 & 2 & Low \\ \hline
\end{tabular}
}
\vspace{-5pt}
\caption{Comparison of base costs to reference tasks. Base Cost per Unit: the cost of annotating one single item (e.g., one QA pair, one paraphrase). All costs are in US cents. *: Additional bonus were rewarded for quality annotators. For example, for our relevance judgements task, we award 1 dime for every 100 quality annotations.}
\vspace{-15pt}
\label{tbl:similarTask}
\end{table*}

\section{Ethical Considerations}
\subsection{Dataset}
\label{ethical_dataset}

\textbf{IRB approval.} All FAQ items were collected in a manner which is consistent with the terms of use of original sources and the intellectual property and privacy rights of the original authors of the texts (i.e., source owners). This project is approved by IRB (institutional review board) at our institution as Exempt Research, which is a human subject study that presents no greater than minimal risk to participants. We consulted data officers at our institution about copyrights. They informed us that ``Website content is generally copyrighted. However, you could claim the concept of \textit{fair use} which allows the use of copyrighted material without permission from the copyright holder when it is used for research, scholarship, and teaching''. We also consulted Section 107\footnote{https://www.copyright.gov/title17/92chap1.html\#107} of U.S. Copyright Act and ensured that our collection action fell under fair use category. We release our dataset under the Creative Commons Attribution-\textit{NonCommercial}-\textit{ShareAlike} 4.0 International License\footnote{https://creativecommons.org/licenses/by-nc-sa/4.0/}.

\noindent \textbf{Annotation via crowdsourcing.} Crowdsourcing involved in this work was conducted on Amazon Mechanical Turk (AMT). In the crowdsourcing step, all participants were required to read and sign an \textit{informed consent form} before participating and they would not be allowed to proceed without signing. AMT mechanism, automatically anonymizing annotators' identities, ensures that the participants’ privacy rights were inherently respected in the crowdsouring process. We determined the \textit{compensation} for each annotation task by evaluating similar tasks on AMT. Table \ref{tbl:similarTask} shows the costs of reference tasks at the time we published our tasks. Overall, taking cognitive complexity into consideration, our base cost per unit is on the same level or higher than reference tasks. Thus, we can safely conclude that crowd workers participating in our annotation tasks were fairly compensated. {Besides, the overall total cost is \$2,683. Considering our competitive base cost per unit and additional generous bonus\footnote{For example, for our relevance judgements task, we award 1 dime for every 100 high-quality annotations.}, we believe that participated annotators are well motivated to contribute high-quality annotations.}

\nop{This project is approved by IRB (institutional review board) and conducted subjected to IRB. This is an Exempt Research, which is a human subject study that presents no greater than minimal risk to participants. Our work is licensed under a Creative Commons Attribution-NonCommercial-ShareAlike 4.0 International License\footnote{https://creativecommons.org/licenses/by-nc-sa/4.0/}. }

\noindent \textbf{Quality check.} {During crowdsourcing phase, we filtered out low-quality annotations. Specifically, we only kept 76.45\% of human-paraphrased queries for the construction of Query Bank by manually checking every single paraphrased query. When constructing the Relevance Set, for each annotator, we sampled a certain number of annotations. If the sampled annotations didn't pass the screening, we dropped all annotations made by that annotator and republished the work again. After such iterative checking, we only kept 89.20\% of annotations in the end.
}

After crowdsourcing, we conducted post-hoc quality checking on both Query Bank and Relevance Set. We manually checked all 1,236 user queries and found that all of them make sense, are related to COVID-19 and properly written. Due to the subjectivity of the relevance judgement task, we evaluated the quality of the relevance annotations in two ways: 1) We find that 73.5\% of $\langle$Query, FAQ item$\rangle$ tuples have high agreements where multiple people vote for the same relevance class; 2) We  re-judge the relevance on randomly sampled 1000 tuples by hiring two research assistants and it turns out that the matching level\footnote{It's considered to be matched if and only if the re-judged score is in the same class (i.e., positive v.s. negative) as the mean of existing annotations.} is 76.5\%. Overall, the post-hoc checking confirms that our \cough dataset is of high quality.

\noindent \textbf{Annotation Protocols.} To further help ethics committees and the public judge the fairness of our annotation process, the annotation protocols for both annotation tasks are listed in Appendix~\ref{protocols}. Figure~\ref{fig:UI_paraphrase} and~\ref{fig:UI_relevance} show the interfaces designed for the annotation process.


\setcounter{table}{0}
\setcounter{figure}{0}
\renewcommand{\thefigure}{A\arabic{figure}}
\renewcommand{\thetable}{A\arabic{table}}

\begin{appendices}
\section{Annotation Protocols}
\label{protocols}
\begin{figure*}[t]
    \centering
    \includegraphics[width=0.8\linewidth]{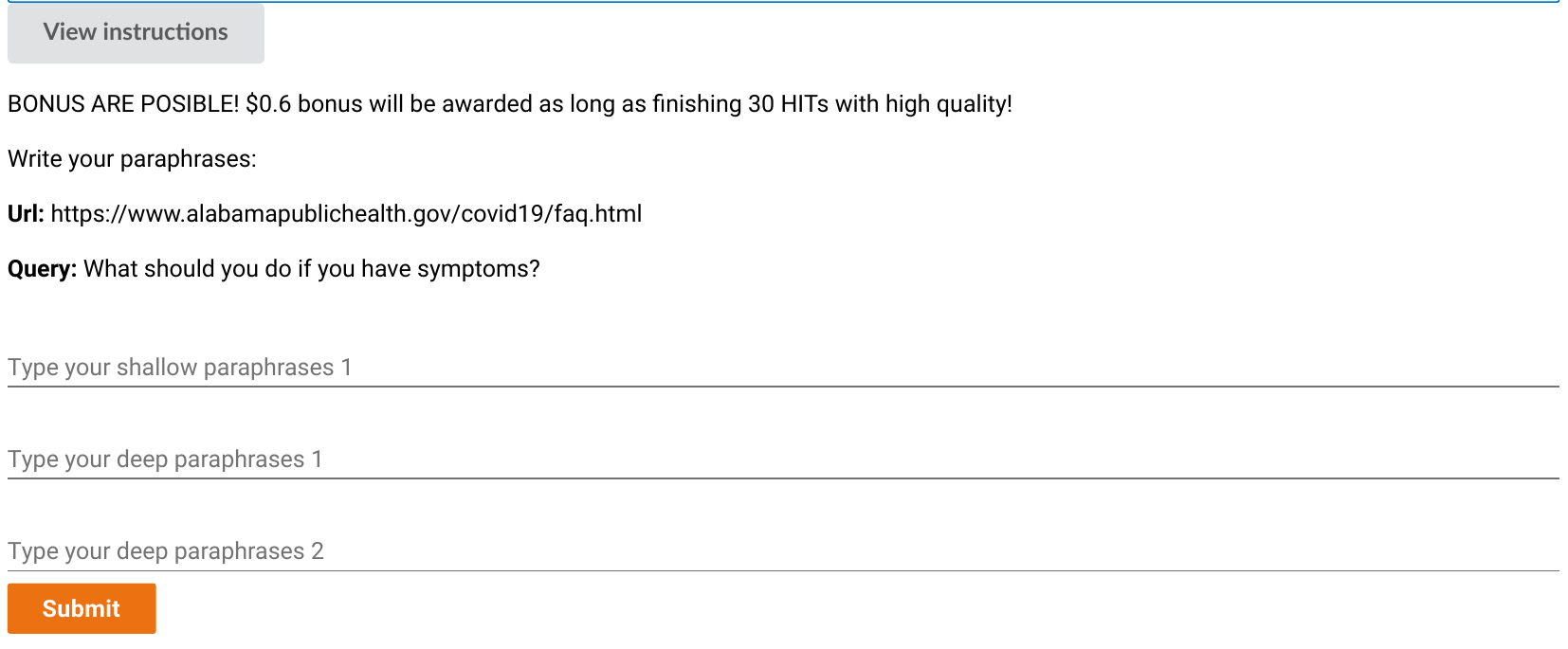}
    \vspace{-5pt}
    \caption{User interface for Query Bank construction task.}
    \label{fig:UI_paraphrase}
    \vspace{-5pt}
\end{figure*}

\begin{figure*}[t]
    \centering
    \includegraphics[width=0.8\linewidth]{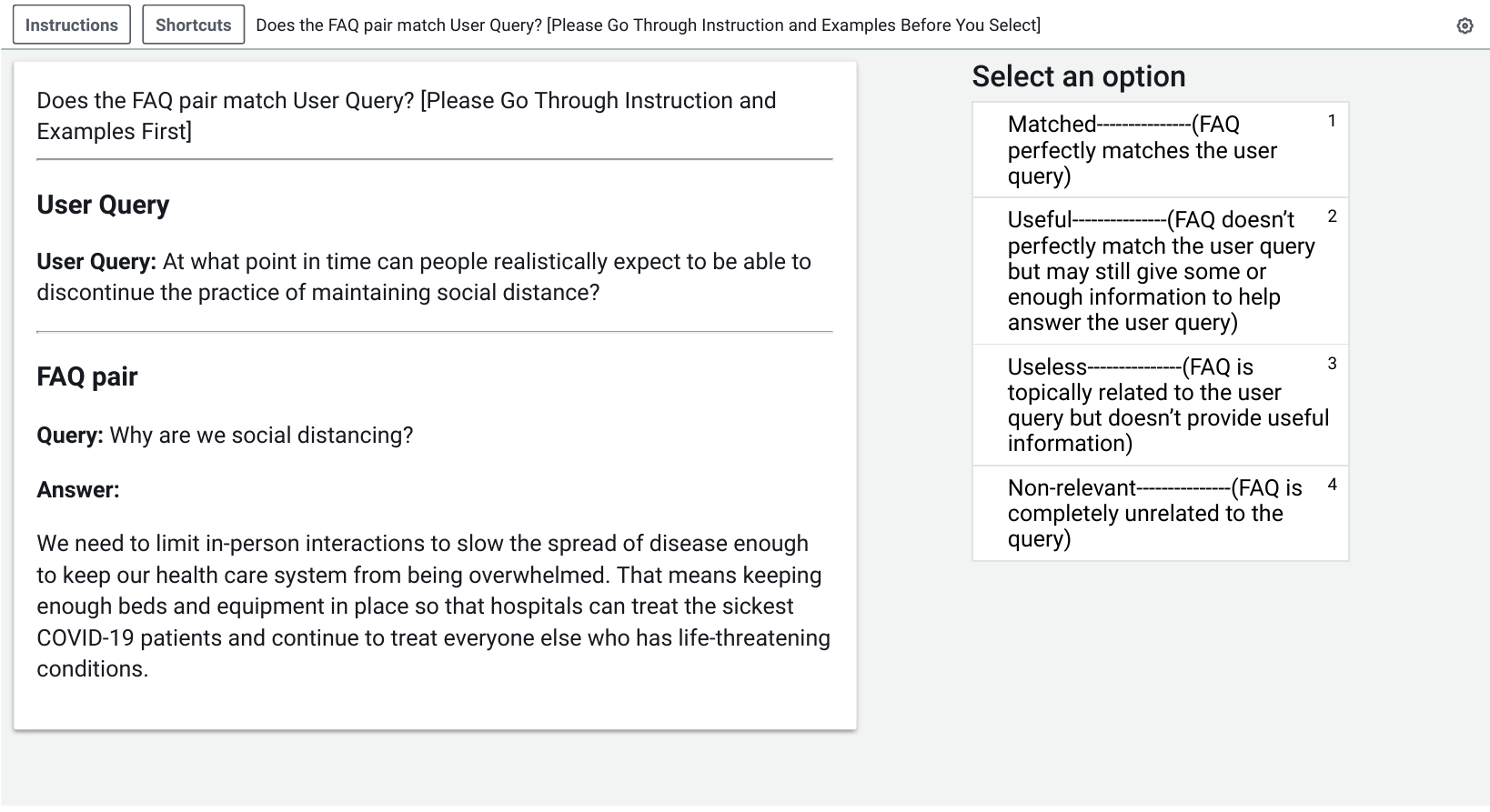}
    \vspace{-5pt}
    \caption{User interface for Relevance Set construction task.}
    \label{fig:UI_relevance}
    \vspace{-5pt}
\end{figure*}

We published our annotation batches on Amazon Mechanical Turk platform. Annotation protocols are provided below to facilitate future research in FAQ retrieval. Figure \ref{fig:UI_paraphrase} and \ref{fig:UI_relevance} show the user interfaces designed for both annotation tasks.

\subsection{Task 1: Query Bank Construction}
For this task, you are expected to give one shallow paraphrase and two deep paraphrases for the query template. Note that query can be either in question form or query string form.

\noindent \textbf{Shallow paraphrase}: Applying word substitution, sentence reordering and other lexical tricks (e.g. extracting salient phrases from response) to the original query to come up with another query without changing the meaning. 

\noindent \textbf{Deep paraphrase}: The paraphrased ones should look dramatically (i.e. grammatically) different from the original query which is more than shallow paraphrasing. However, the paraphrased query should share the same (or almost same) semantic meaning as the original query.

\subsection{Task 2: Relevance Set Construction}
\label{task2annotation}
For this task, you will see a FAQ item retrieved by an automatic tool for a particular user query, and your job is to judge the relevance of the FAQ item based on the annotation scheme shown below.

    \noindent \textbf{Matched}: The candidate FAQ  matches the user query perfectly. (Query part of FAQ is semantically identical to the user query, and answer part of FAQ well answers the user query.)
    
    \noindent \textbf{Useful}: The candidate FAQ doesn’t perfectly match the user query but may still give some or enough information to help answer the user query. (Query part of FAQ is semantically similar to the user query, and you can either extract or infer some information from the answer which could be useful to the user query. Or alternatively, the candidate FAQ provides too much extra information which is not necessary.)
    
    \noindent \textbf{Useless}: The candidate FAQ is topically related to the user query but doesn’t provide useful information. (Query part of FAQ is somewhat related to the user query, but you can’t get any useful information out of the answer part to confidently answer the user query.)
    
    \noindent \textbf{Non-relevant}: The candidate FAQ is completely unrelated to the query.

\begin{table}[t]
\centering
\resizebox{0.85\linewidth}{!}{
\begin{tabular}{lrrrrr}
\hline
Method & P@1 & P@5 & MAP & MRR & nDCG \\ \hline
BM25 (Q-q) & 43.3 & 26.8 & 25.0 & 57.0 & 76.7 \\
BM25 (Q-a) & 21.0 & 15.6 & 14.1 & 33.9 & 46.4 \\
BM25 (Q-q+a) & 37.6 & 25.2 & 24.4 & 52.4 & 72.6 \\ \hline
BERT (Q-q) & 46.0 & 28.4 & 24.8 & 59.2 & \textbf{78.6} \\
\, + fine-tune on pesudo Q-q & 49.9 & 27.2 & 26.5 & 61.1 & 63.0 \\
BERT (Q-a) & 8.3 & 5.9 & 4.7 & 16.3 & 16.7 \\
\, + fine-tune on FAQ Bank & 35.4 & 23.6 & 22.8 & 49.8 & 56.4 \\
\, \, \, + re-rank & 35.6 & 24.2 & 23.4 & 50.6 & 57.8 \\ \hline
CombSum & \textbf{51.6} & \textbf{31.2} & \textbf{32.6} & \textbf{64.8} & 74.7 \\
\, - fine-tuned BERT (Q-a) & 47.8 & 29.0 & 28.1 & 61.6 & 75.2 \\
\hline
\end{tabular}
}
\vspace{-5pt}
\caption{Evaluation on \cough (Aggregation scheme B).}

\label{tbl:extra1}
\end{table}
\begin{table}[t]
\centering
\resizebox{0.85\linewidth}{!}{
\begin{tabular}{lrrrrr}
\hline
Method & P@1 & P@5 & MAP & MRR & nDCG \\ \hline
BM25 (Q-q) & 66.0 & 50.2 & 28.5 & 77.5 & 76.7 \\
BM25 (Q-a) & 38.4 & 29.2 & 15.6 & 52.5 & 46.4 \\
BM25 (Q-q+a) & 61.3 & 47.2 & 27.8 & 74.5 & 72.6 \\ \hline
BERT (Q-q) & 70.4 & \textbf{53.5} & 28.2 & 80.9 & \textbf{78.6} \\
\, + fine-tune on pesudo Q-q & 70.7 & 44.9 & 26.0 & 79.7 & 63.0 \\
BERT (Q-a) & 15.7 & 11.0 & 4.8 & 26.7 & 16.7 \\
\, + fine-tune on FAQ Bank & 55.1 & 39.8 & 24.0 & 68.7 & 56.4 \\
\, \, \, + re-rank & 54.6 & 40.9 & 24.4 & 68.6 & 57.8 \\ \hline
CombSum & \textbf{72.3} & 52.9 & \textbf{35.8} & \textbf{82.4} & 74.7 \\
\, - fine-tuned BERT (Q-a) & 70.2 & 51.1 & 31.3 & 80.9 & 75.2
\\ \hline
\end{tabular}
}
\vspace{-5pt}
\caption{Evaluation on \cough (Aggregation scheme C).}

\label{tbl:extra2}
\end{table}
\begin{table}[t]
\centering
\resizebox{0.85\linewidth}{!}{
\begin{tabular}{lrrrrr}
\hline
Method & P@1 & P@5 & MAP & MRR & nDCG \\ \hline
BM25 (Q-q) & 77.1 & 65.8 & 32.2 & 86.1 & 76.7 \\
BM25 (Q-a) & 49.5 & 39.2 & 18.3 & 62.4 & 46.4 \\
BM25 (Q-q+a) & 76.0 & 62.8 & 32.7 & 85.2 & 72.6 \\ \hline
BERT (Q-q) & 81.6 & \textbf{68.5} & 30.7 & 89.1 & \textbf{78.6} \\
\, + fine-tune on pesudo Q-q & 77.5 & 54.9 & 27.3 & 85.0 & 63.0 \\
BERT (Q-a) & 20.5 & 14.1 & 5.1 & 32.4 & 16.7 \\
\, + fine-tune on FAQ Bank & 66.8 & 51.2 & 27.1 & 78.1 & 56.4 \\
\, \, \, + re-rank & 67.2 & 52.9 & 27.6 & 78.4 & 57.8 \\ \hline
CombSum & \textbf{84.3} & 67.4 & \textbf{40.8} & \textbf{90.6} & 74.7 \\
\, - fine-tuned BERT (Q-a) & 80.9 & 65.5 & 35.1 & 88.5 & 75.2 
\\ \hline
\end{tabular}
}
\vspace{-5pt}
\caption{Evaluation on \cough (Aggregation scheme D).}

\label{tbl:extra3}
\end{table}

\section{Aggregation Schemes}
\label{labeling}
In this work, we introduce four aggregation schemes to obtain binary labels.
\begin{itemize}[leftmargin=1em,noitemsep,topsep=0pt,parsep=0pt,partopsep=0pt]
    \item[A.] Annotated $\langle$Query, FAQ item$\rangle$ tuples with mean score $\geq$ 3 are positives. 
    \item[B.] Annotated $\langle$Query, FAQ item$\rangle$ tuples with mean score $>$ 3 are positives. 
    \item[C.] Annotated $\langle$Query, FAQ item$\rangle$ tuples that have at least one\footnote{For tuples with more than 3 annotations, we raise the bar to two ``Matched''.} ``Matched'' annotation are positives. 
    \item[D.] For each annotated $\langle$Query, FAQ item$\rangle$ tuple, we convert ``Matched'' and ``Useful'' to positive annotations, and ``Useless'' and ``Non-relevant'' to negative annotations. We then apply majority voting using converted binary annotations.
\end{itemize}

\subsection{Results for Different Aggregation Schemes}
\label{labeling_results}
Results based on aggregation schemes B, C and D are shown in Tables~\ref{tbl:extra1}, \ref{tbl:extra2} and \ref{tbl:extra3}, respectively. Results based on aggregation scheme A are shown in Table~\ref{tbl:extra}.

\section{Implementation Details}
We first preprocess user query and FAQ items with nltk porter stemmer 5\footnote{https://www.nltk.org/}. For baselines including BM25\footnote{https://pypi.org/project/rank-bm25/} and Sentence-BERT\footnote{https://github.com/UKPLab/sentence-transformers and we use \textit{distilbert-base-nli-stsb-quora-ranking} model card.}, we take the standard off-the-shelf version. More specifically, we keep the default $k1$ as 2 and  $b$ as $0.75$ for BM25 over Q-q, Q-a and Q-q+a settings. When deploying synthetic query generation model (i.e., GPT2), hyper-parameters are set as instructed by \citet{Unsupervised-FAQ} (see their Section 3.4). We adopt the in-batch negatives training strategy to fine-tune both Sentence-BERT and cross-encoder BERT. For both BERT models, we use the Adam optimizer \cite{KingmaB14} with a learning rate of 1e-5 and fine-tune up to 10 epochs. We set the batch sizes as 24 and 4 for Sentence-BERT and cross-encoder BERT, respectively. All experiments are conducted using one single GeForce GTX 2080 Ti 12 GB GPU (with significant CPU resources).

\clearpage
\begin{table}[t!]
\resizebox{\linewidth}{!}{%
\begin{tabular}{lc}
\hline
 & \# of FAQ Items \\ \hline
Arizona Health Care Cost Containment System&138 \\ 
Alabama Public Health&89 \\ 
American Medical Association&14 \\ 
California Department of Health&28 \\ 
Government of Canada&131 \\ 
Centers for Disease Control and Prevention&378 \\ 
Children's Hospital Los Angeles&73 \\ 
Bloomberg Harvard City Leadership Initiative&186 \\ 
Cleveland Clinic&15 \\ 
CNN&112 \\ 
Government of Colorado&66 \\ 
Delaware Department of Health&71 \\ 
U.S. Food and Drug Administration&139 \\ 
European Centre for Disease Prevention and Control&55 \\ 
Florida Department of Health&47 \\ 
Georgia Department of Labor&16 \\ 
Explore Georgia&13 \\ 
Government of United Kingdom&53 \\ 
Harvard Health Publishing&104 \\ 
Illinois Department of Public Health&37 \\ 
Inspire&1753 \\ 
JHU HUB&7 \\ 
JHU Medicine&14 \\ 
Kids Health from Nemours&121 \\ 
King County, Washington&26 \\ 
Government of Massachusetts&17 \\ 
Medical News Today&28 \\ 
MedHelp&282 \\ 
Government of Michigan&75 \\ 
Minnesota Department of Health&98 \\ 
New York Times&100 \\ 
Government of New Jersey&322 \\ 
National Institute of Health&105 \\ 
Government of North Carolina&59 \\ 
Government of New York&75 \\ 
New York State Electric and Gas&68 \\ 
New York Department of Financial Services&45 \\ 
Pennsylvania Office of Unemployment Compensation&222 \\ 
Government of Pennsylvania&66 \\ 
University of Pennsylvania Health System&63 \\ 
Sante Clara Department of Health&103 \\ 
San Mateo County Health&47 \\ 
Texas Health Services&39 \\ 
Tricare&94 \\ 
United Nations&40 \\ 
United States Department of Agriculture&152 \\ 
United States Department of Labor&43 \\ 
Virginia Department of Health&435 \\ 
United States Department of Veterans Affairs&16 \\ 
Washington Department of Health&137 \\ 
WHO&29 \\ 
World Health Organization&395 \\ 
WikiHow&2371 \\

 \hline
Total & 9151  \\ 
\end{tabular}
}
\caption{Number of English FAQ items scrapped from each source.}
\vspace{-15pt}
\label{tbl:index_English}
\end{table}
\begin{table}[t!]
\resizebox{\linewidth}{!}{%
\begin{tabular}{lcc}
\hline
 & language & \# of FAQ Items \\ \hline
Centers for Disease Control and Prevention&Spanish&268 \\ 
Centers for Disease Control and Prevention&Korean&244 \\ 
Centers for Disease Control and Prevention&Vietnamese&244 \\ 
Centers for Disease Control and Prevention&Chinese&244 \\ 
Children's Hospital Los Angeles&Arabic&30 \\ 
Children's Hospital Los Angeles&Spanish&45 \\ 
Children's Hospital Los Angeles&Persian&39 \\ 
Children's Hospital Los Angeles&Armenian&38 \\ 
Children's Hospital Los Angeles&Kanuri&32 \\ 
Children's Hospital Los Angeles&Chinese&34 \\ 
U.S. Food and Drug Administration&Spanish&83 \\ 
Japan Health&Japanese&226 \\ 
Japan Labor&Japanese&63 \\ 
United Nations&Arabic&39 \\ 
United Nations&Spanish&38 \\ 
United Nations&French&37 \\ 
United Nations&Chinese&38 \\ 
World Health Organization&Arabic&328 \\ 
World Health Organization&Spanish&356 \\ 
World Health Organization&French&387 \\ 
World Health Organization&Russian&301 \\ 
World Health Organization&Chinese&367 \\ 
WikiHow&Arabic&144 \\ 
WikiHow&Czech&22 \\ 
WikiHow&German&525 \\ 
WikiHow&Spanish&310 \\ 
WikiHow&Persian&49 \\ 
WikiHow&French&301 \\ 
WikiHow&Hindi&286 \\ 
WikiHow&Indonesian&166 \\ 
WikiHow&Italian&263 \\ 
WikiHow&Japanese&286 \\ 
WikiHow&Korean&128 \\ 
WikiHow&Dutch&142 \\ 
WikiHow&Portuguese&303 \\ 
WikiHow&Russian&142 \\ 
WikiHow&Thai&90 \\ 
WikiHow&Vietnamese&101 \\ 
WikiHow&Chinese&117 \\

  \hline
Total & - & 6768
\end{tabular}%
}
\caption{Number of non-English FAQ items scrapped from each source and language.}
\vspace{-15pt}
\label{tbl:index_Non-English}
\end{table}
\end{appendices}

\end{document}